\documentclass[sigconf]{acmart}
\usepackage[linesnumbered, ruled]{algorithm2e}
\SetKwRepeat{Do}{do}{while}%

\AtBeginDocument{%
  \providecommand\BibTeX{{%
    \normalfont B\kern-0.5em{\scshape i\kern-0.25em b}\kern-0.8em\TeX}}}

\setcopyright{none}
\copyrightyear{2021}
\acmYear{2021}
\acmDOI{10.1145/1122445.1122456}
\renewcommand\footnotetextcopyrightpermission[1]{}
\begin{document}

% \title{T2IVAE: Time Series to Image VAE for Unsupervised Time Series Anomaly Detection}
\title{NVAE-GAN Based Approach for Unsupervised Time Series Anomaly Detection}

\author{Liang Xu}
\authornote{Both authors contributed equally to this research.}
\email{xlpaul@126.com}
\author{Liying Zheng}
\authornotemark[1]
\email{liyingzheng33@gmail.com}
\affiliation{%
  % \institution{Ping An Technology}
}


\author{Weijun Li}
\email{liweijun382@gmail.com}
\affiliation{%
  % \institution{Ping An Technology}
  }

\author{Zhenbo Chen}
\email{zhenbochen92@gmail.com}
\affiliation{%
  %\institution{Ping An Technology}
  }

\author{Weishun Song}
\email{songweishun474@gmail.com}
\affiliation{%
  % \institution{Ping An Technology}
  }
  
\author{Yue Deng}
\email{dengyue606@gmail.com}
\affiliation{%
  % \institution{Ping An Technology}
  }

\author{Yongzhe Chang}
\email{changyongzhe@sz.tsinghua.edu.cn}
\affiliation{%
  % \institution{Tsinghua University}
  }

\author{Jing Xiao}
\authornote{Jing Xiao and Bo Yuan are corresponding authors.}
\email{xiaojing661@pingan.com.cn}
\affiliation{%
  % \institution{Ping An Technology}
  }

\author{Bo Yuan}
\authornotemark[2]
\email{yuanb@sz.tsinghua.edu.cn}
\affiliation{%
  % \institution{Tsinghua University}
  }

% \affiliation{%

% \affiliation{%

% \affiliation{%

% \affiliation{%

% \affiliation{%

% \affiliation{%

\renewcommand{\shortauthors}{ }

% \begin{abstract}
%  Time series anomaly detection is a very common task in many industries, such as network monitoring, facility maintenance, and information security. However, it is still a challenging task to detect abnormal samples in time series with high accuracy, due to noisy data collected from real world, and multiple complicated abnormal patterns to capture. In this paper, we focus on unsupervised learning algorithms, inspired by Variational Auto-Encoder (VAE), which is a kind of reconstruction-based deep learning model and is frequently used for time series anomaly detection in recent years. We propose Time 
% series to Image VAE (T2IVAE), an unsupervised model for univariate series. T2IVAE takes 1D time series as input and transforms it to 2D image, and then adopts the reconstruction loss to detect abnormal time steps in original series. Furthermore, we apply the techniques in Generative Adversarial Networks (GAN) to T2IVAE training, trying to improve model robustness through generated samples. To validate our model, we have made experiments on three datasets, and compared it with other several popular models using F1 score. T2IVAE achieves 0.504 on our dataset collected from real-world scenario, 0.639 on Numenta datasets, and 0.651 on NASA datasets, which makes it outperform other comparison models.
% \end{abstract}

\begin{abstract}
In recent studies, Lots of work has been done to solve time series anomaly detection by applying Variational Auto-Encoders (VAEs). Time series anomaly detection is a very common but challenging task in many industries, which plays an important role in network monitoring, facility maintenance, information security, and so on. However, it is very difficult to detect anomalies in time series with high accuracy, due to noisy data collected from real world, and complicated abnormal patterns. From recent studies, we are inspired by Nouveau VAE (NVAE) and propose our anomaly detection model: Time series to Image VAE (T2IVAE), an unsupervised model based on NVAE for univariate series, transforming 1D time series to 2D image as input, and adopting the reconstruction error to detect anomalies. Besides, we also apply the Generative Adversarial Networks based techniques to T2IVAE training strategy, aiming to reduce  the overfitting. We evaluate our model performance on three datasets, and compare it with other several popular models using F1 score. T2IVAE achieves 0.639 on Numenta Anomaly Benchmark, 0.651 on public dataset from NASA, and 0.504 on our dataset collected from real-world scenario, outperforms other comparison models.
\end{abstract}





\maketitle

\section{Introduction}
Time series data exist in many real-world scenarios, such as network monitoring, facility maintenance, and information security. Anomaly detection for time series is a common but important task in many industries. For example, anomalies in the website visiting volume might reflect hacker attacks or access failures, and anomalies in facility sensor data might indicate damage or malfunction. Due to the commonality and importance, anomaly detection algorithms with high performance for time series are essential and indispensable for may related industries. Current research on time series analysis can be classified into two main categories, univariate analysis and multivariate analysis, where univariate analysis focus on single time sequence, and multivariate analysis focus on simultaneous multiple time sequences. In this paper, we focus on univariate anomaly detection.

Same as many former models, our model takes time series with multiple time steps as input, and detect time steps with anomaly values. To make it clear, we have time series input
\begin{math}
  S
\end{math}
with
\begin{math}
  L
\end{math}
time steps:
\begin{displaymath}
  S=\left \{s_{1},s_{2},\cdots ,s_{l-1},s_{l} \right \}
\end{displaymath}
and 
\begin{math}
  s_{t}
\end{math}
indicates the collected single value at time step
\begin{math}
  t
\end{math}. Our model will give results on which are anomalies in these
\begin{math}
  L
\end{math}
time steps. This is a very typical task in data mining, and lots of statistical models and machine learning models are designed for this task, such as S-H-ESD\cite{hochenbaum2017automatic}, One-Class SVM\cite{ma2003time}, etc. However, time series collected from real world are usually very complicated, making these techniques not suitable for all kinds of anomaly patterns, due to their limitations. For example, S-H-ESD can be only used in time series with periodicity, and One-Class SVM relies heavily on feature engineering to gain good performance. Deep learning models are able to capture more complicated hidden features and temporal correlations, which can help us detect different kinds of anomalies. Even so, there still exists some problems to mention for applying deep learning.

The first problem is lacking of labeled data with lots of abnormal samples required for normal training phase, as supervised learning is an intuitive method to get a detection model if labeled data is provided, like RobustTAD\cite{gao2020robusttad} and SRCNN\cite{ren2019time}. However, labeled data is quite limited and imbalanced in time series scenarios, especially in real world scenarios. Thus we need to design the model not to rely heavily on labeled data, in this way we can extend the applicability of our model in many real world scenarios.

Second problem is that, real-world time series contain noises and multiple complicated abnormal patterns, which makes it very tough to distinguish between normal samples and abnormal samples, or to extract effective hidden features to achieve higher robustness. Deep learning algorithms with shallow structures are not efficient enough to deal with complex temporal features and correlations. In lots of recent studies, Variational Auto-Encoder\cite{kingma2013auto} (VAE) based models have been frequently used and analyzed for time series anomaly detection\cite{xu2018unsupervised}\cite{su2019robust}, and have been proved to perform well in this scenario.

Considering the mentioned problems above, we propose Time series to Image VAE (T2IVAE), an unsupervised model based on NVAE\cite{vahdat2020nvae} for time series detection. T2IVAE is a reconstruction-based deep learning model as most VAE models do, trying to reconstruct the given input and then generate the output, and finally calculating anomaly score based on the defined reconstruction error to find abnormal time steps. There are three key improvements in T2IVAE: (1) Transform input time series to 2D image before input, making the input contain more temporal features and correlations. (2) Apply NVAE to time series problem, with deep hierarchical VAE structure to improve time series reconstruction. (3) Apply Generative Adversarial Networks\cite{goodfellow2014generative}\cite{huang2018introvae} (GAN) in training phase to reduce overfitting.

To evaluate T2IVAE on time series anomaly detection and compare it with other models, we have made experiments on three datasets, namely dataset of network switches collected from our own real-world scenario, public dataset Numenta, and public dataset NASA. Results show that T2IVAE achieves 0.504 F1 score on our data, 0.639 F1 score on Numenta\cite{lavin2015evaluating}\footnote[1]{The data is available at: https://github.com/numenta/NAB}, and 0.651 F1 score on NASA\footnote[2]{The data is available at: https://s3-us-west-2.amazonaws.com/
telemanom/data.zip}.

\section{Related Work}
Recently, a lot of research work based on deep learning have been developed to solve unsupervised anomaly detection problems, VAE and GAN are two most frequently adopted frameworks, as DONUT and TADGAN, which proves the effectiveness of reconstruction-based methods.

\textbf{DONUT: }
DONUT\cite{xu2018unsupervised} is a relatively early but famous research to build unsupervised anomaly detection algorithm based on VAE, trying to detect anomalies through reconstruction probability. To adjust model applicable for time series, several improvements have been made including modified ELBO loss, missing data injection, and MCMC imputation for detection.

\textbf{TADGAN: }
TADGAN\cite{geiger2020tadgan} is a recent unsupervised anomaly detection model based on GAN. TADGAN uses LSTM layers as base models for generators and critics, trying to capture the temporal correlations of time series distributions. Furthermore, it has made a lot of effort to adapt this algorithm to time series, including training with cycle consistency loss to allow
for effective time-series data reconstruction, and proposing several approaches to compute anomaly scores. TADGAN has been evaluated on several public datasets with other baseline methods, and the results show that this model outperforms
baseline methods with the highest averaged F1 score across all the datasets.

Meanwhile, we have noticed NVAE from NVIDIA, has greatly improved VAE-based image generation by hierarchical VAE structure, and achieved state-of-the-art results on several datasets. Such progress means there’s still space for improvements in VAE-based models, maybe also we can benefit from it for time series anomaly detection scenario. Inspired by this, we have an idea that transfer time series into images for NVAE-based detection model. Furthermore, through transferring time series into images, we are able to include more complicated temporal correlation between time steps as input for detection model.

\textbf{NVAE: }
NVAE\cite{vahdat2020nvae}, a deep hierarchical VAE built for image generation using depth-wise separable convolutions and residual parameterization of Normal distributions. Such designed network structure helps NVAE achieve state-of-the-art results among non-autoregressive likelihood-based models on several public datasets, such as MNIST and CIFAR-10.

\textbf{Sequence 2D Encoding: }
For encoding time sequences into images, we’ve investigated Recurrence Plots\cite{eckmann1995recurrence}, and Gramian Angular Field (GAF)\cite{wang2015encoding}. Suppose Given a univariate time series of length N, these algorithms will encode it and output a N×N image. Recurrence Plots calculates distance between any two sub-trajectories of a time sequence, then build the image. GAF turns sequence into polar coordinates and calculate trigonometric sum measuring the temporal correlation within different time intervals, and this algorithm has already been used with convolution classifier for time series classification.

Considering these previous work, our main idea is to follow the VAE-based detection method, but encode input time sequences into 2D images at first and adopt NVAE as the core part of model, trying to improve the performance on time series anomaly detection.

\section{T2IVAE}

\begin{figure}[h]
  \centering
  \includegraphics[width=\linewidth]{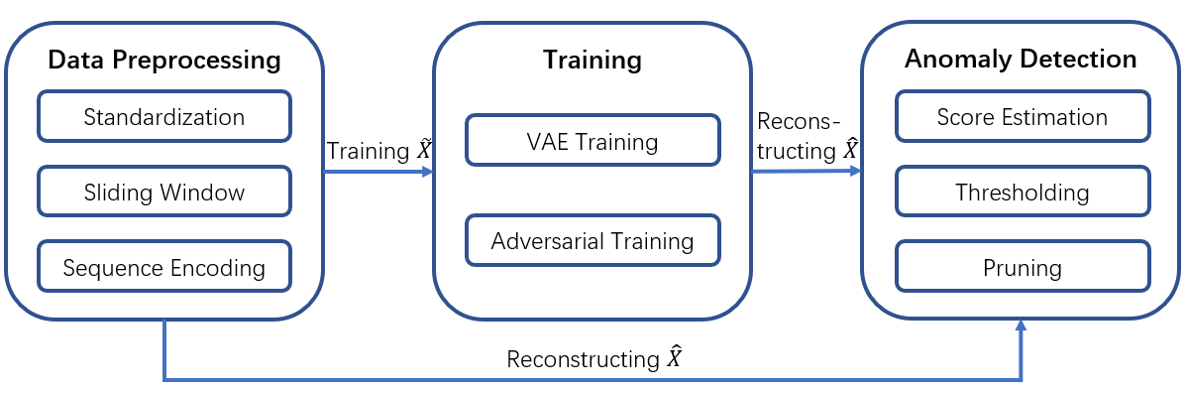}
  \caption{Overall architecture of T2IVAE.}
  \label{Overall architecture of T2IVAE.}
  \Description{Overall architecture of T2IVAE.}
\end{figure}

We propose Time series to Image VAE (T2IVAE), unsupervised time series anomaly detection model based on VAE. The overall architecture of T2IVAE is shown as Fig.(\ref{Overall architecture of T2IVAE.}). Sequence Encoding, model structure, training strategy and detection strategy are T2IVAE's key components, which are going to be introduced in detail in this section.

\subsection{Sequence 2D Encoding}
Current anomaly detection models get 1D time series as input. To make it clear, we have a time series input
\begin{math}
  S
\end{math}
with
\begin{math}
  L
\end{math}
time steps:
\begin{displaymath}
  S=\left \{s_{1},s_{2},\cdots ,s_{l-1},s_{l} \right \}
\end{displaymath}
and 
\begin{math}
  s_{t}
\end{math}
indicates the value at time step
\begin{math}
  t
\end{math}.
As mentioned above, our idea is to encode 1D time series into 2D images (in order to capture more information of time series correlation), then we can apply convolution VAE model which can perform well on 2D feature extracting. To predict a specific value at time step
\begin{math}
  k
\end{math}
is abnormal or not, we extract a sub sequence with window size
\begin{math}
  N=2\times w
\end{math}
around time step
\begin{math}
  k
\end{math}
, and get sub-sequence 
\begin{math}
  X_{k}
\end{math}
:
\begin{displaymath}
  X_{k}=\left \{s_{k-w+1},\cdots ,s_{k},\cdots,s_{k+w} \right \}
\end{displaymath}

After encoding, time series X will be transferred to a new matrix with shape as
\begin{math}
\left [N,N \right ]
\end{math}
, so the hidden states of our proposed model will have shape like
\begin{math}
\left [B,N,N,C \right ]
\end{math}
(
\begin{math}
  B
\end{math}
indicates batch and
\begin{math}
  C
\end{math}
indicates amount of channels). In our experiments, we use Gramian Angular Field (GAF) and Recurrence Plots for encoding, so we will have 2 encoded matrix and concatenate them to get
\begin{math}
\check{X}
\end{math}
as input for NVAE network with shape as
\begin{math}
\left [B,N,N,2 \right ]
\end{math}
.

\textbf{Gramian Angular Field: }
GAF\cite{wang2015encoding}\footnote[3]{Code implementation is available at: https://github.com/johannfaouzi/pyts} normalize observations of a time series into interval of
\begin{math}
\left [-1,1 \right ]
\end{math}
and get 
\begin{math}
\bar{X}_{k}
\end{math}
:
\begin{equation}
  \bar{x}_{k,i}=\frac{x_{i}-max\left ( X_{k}\right )+\left (x_{i}-min\left (X_{k}\right ) \right )}{max\left ( X_{k}\right )-min\left ( X_{k}\right )},  \forall i \in \{1, \cdots, n\}
\end{equation}

\begin{figure}[h]
  \centering
  \includegraphics[width=\linewidth]{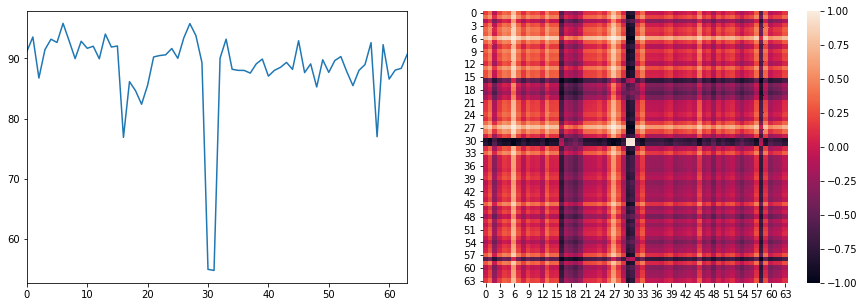}
  \caption{Origin series and encoded image by GAF, for a sub-sequence of time series with abnormal value below 60 from Numenta dataset.}
  \label{gaf}
  \Description{Encoding figure of GAF.}
\end{figure}

Then GAF will transfer
\begin{math}
\bar{X}_{k}
\end{math}
into polar coordinates by encoding it as angular cosine, then calculate the trigonometric sum between each time step to identify the temporal correlation within different time intervals. For time series input X, we can calculate
\begin{math}
X^{GAF}_{k}
\end{math}
as:
\begin{equation}
X^{GAF}_{k}=(\bar{X}_{k})^{T}\cdot \bar{X}_{k}-\sqrt{I-(\bar{X}_{k})^{2}}^{T}\cdot \sqrt{I-(\bar{X}_{k})^{2}}
\end{equation}
and 
\begin{math}
I
\end{math}
is the unit row vector
\begin{math}
\left [1,1,\cdots ,1 \right ]
\end{math}
. Normal values are close to nearby ones, thus also close in polar coordinates. However, when abnormal values exist, such as outliers or change points, then there will be significant difference between normal parts and abnormal parts in an image just like anomalies on images, since the trigonometric sum calculates for every 2 time steps and more sensitively to capture anomalies. One example is shown in Fig.(\ref{gaf}), a sequence containing outliers after GAF encoding. This property make it possible to use model with 2D input to detect anomalies in 1D time series.

\textbf{Recurrence Plots: }
A recurrence plot is an image representing the distances between trajectories extracted from the original time series\cite{eckmann1995recurrence}\footnotemark[3]. For time series input X, we can extract trajectories and get 
\begin{math}
\vec{X}_{k}
\end{math}
as:
\begin{equation}
\vec{x}_{k,i} = (x_{k,i}, x_{k,i+\tau}, \cdots, x_{k,i+(m-1)\tau}), \quad \forall i \in \{1, \cdots, n-(m-1)\tau \}
\end{equation}
here 
\begin{math}
\vec{m}
\end{math}
indicates the dimension of trajectories, and
\begin{math}
\tau
\end{math}
indicates time delay. After that we can get encoded 
\begin{math}
X^{REC}_{k}
\end{math}
as:
\begin{equation}
x^{REC}_{k,i, j} = \| \vec{x}_{k,i}-\vec{x}_{k,j} \|, \quad \forall i,j \in \{1, \cdots, n-(m-1)\tau \}
\end{equation}
In our experiments, we set both
\begin{math}
\vec{m}
\end{math}
and
\begin{math}
\tau
\end{math}
to 1, so make it as a point-wise distance calculation.

\begin{figure}[h]
  \centering
  \includegraphics[width=\linewidth]{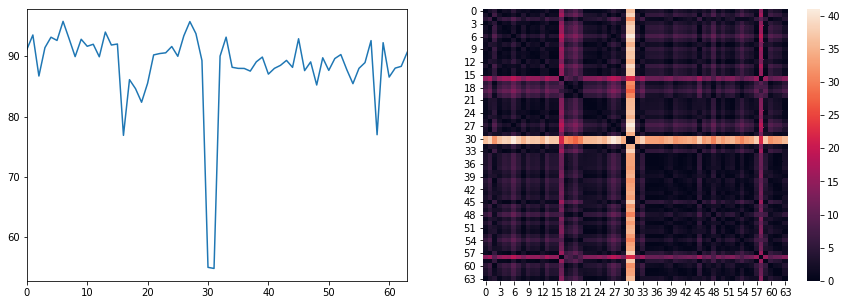}
  \caption{Origin series and encoded image by Recurrence Plots, for a sub-sequence of time series with abnormal value below 60 from Numenta dataset.}
  \label{rec}
  \Description{Encoding figure of Recurrence Plots.}
\end{figure}

When an abnormal time step exist, distances with other normal ones will be much larger, then there will be several anomaly distances on the image, that will make our model to detect. One example is shown in Fig.(\ref{rec}), a sequence containing outliers after Recurrence Plots encoding.

After encoding by GAF and Recurrence Plots, we can concatenate them along channel axis and get
\begin{math}
\check{X}_{k}
\end{math}
as input of NVAE networks:
\begin{equation}
\check{X}_{k}=Concatenate(X^{GAF}_{k},X^{REC}_{k})
\end{equation}
Since we have multiple time steps for a time series, then we can have set
\begin{math}
\check{X}
\end{math}
including all time steps for anomaly detection.

\subsection{Modeling}

After the preprocessing of the time series input
\begin{math}
  S
\end{math}, including standardization, sliding windows and sequence encoding, we get a set of tensor 
\begin{math}
\check{X}
\end{math} for all time steps
, which is then batch processing into \begin{math}
\tilde{X}
\end{math} with batch size \begin{math}
B
\end{math} and input to T2IVAE for sample encoding and reconstruction decoding.

In T2IVAE, a hierarchical multi-scale encoder-decoder architecture is adopted, modeling prior distribution of latent variables
\begin{math}
  q(z)
\end{math} and posterior distribution of latent variables
\begin{math}
  p(z|\tilde{X})
\end{math} with auto-regressive Gaussian model. Specifically, the latent variables is partitioned into disjoint groups, 
\begin{displaymath}
  z=\left \{z_{1},z_{2},\cdots ,z_{G-1},z_{G} \right \},
\end{displaymath} where 
\begin{math}
  G
\end{math} is the number of groups and 
\begin{math}
  z_{g}
\end{math} is a latent sub-variables. Then, we have 
\begin{equation}
  q(z)=\prod_{g=1}^{G}q(z_{g}|z_{<g}),
\end{equation}
\begin{equation}
  p(z|\tilde{X})=\prod_{g=1}^{G}p(z_{g}|z_{<g},\tilde{X}).
\end{equation}
Each group of 
\begin{math}
  q(z)
\end{math} is assumed as factorial Normal distributions. On the encoder side, we first extract representation from input
\begin{math}
  \tilde{X}
\end{math} with a bottom-up deterministic network, then auto-regressively sample 
\begin{math}
  z_{g}
\end{math} with a top-down network. On the decoder side, the top-down network of encoder is reused to auto-regressively sample 
\begin{math}
  z_{g}
\end{math} and finally reconstruct 
\begin{math}
  \hat{X}
\end{math}.

The setting of residual cells of the T2IVAE encoder and decoder is the same as that of NVAE. The main function of residual cells is to improve the modeling ability of data with long-term correlation by increasing the reception field of the network with multi-layer convolution.

\begin{figure}[h]\label{structure}
  \centering
  \includegraphics[width=0.9\linewidth]{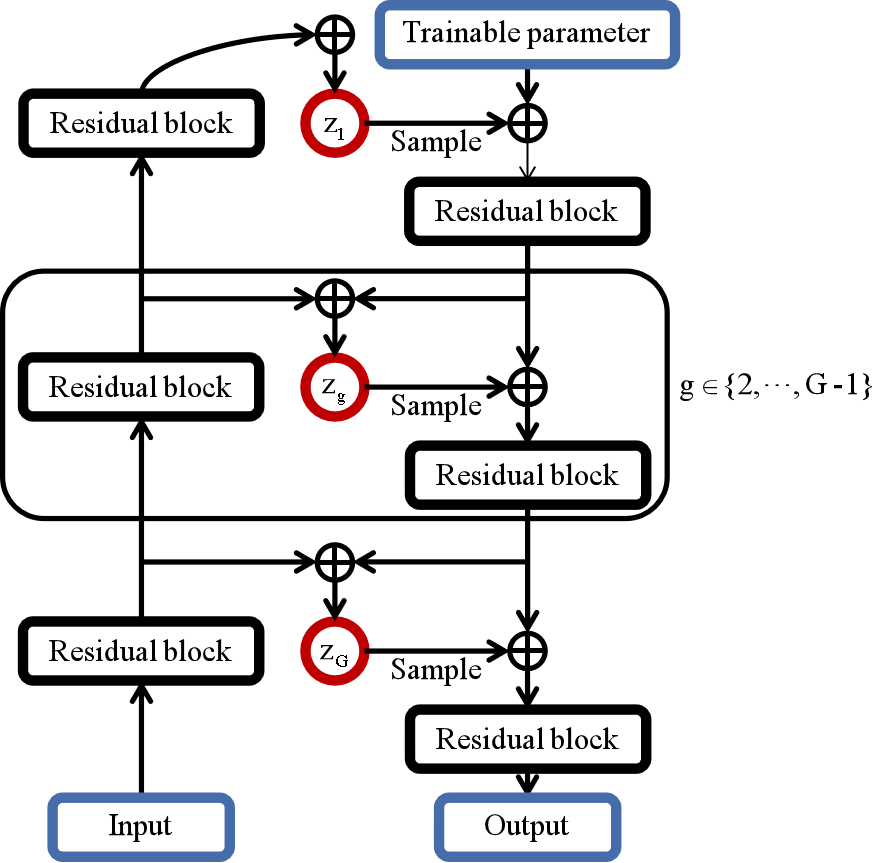}
  \caption{Structure of T2IVAE, with G groups of hierarchical VAE. Residual block denotes residual neural networks.}
  \label{Structure draft}
  \Description{Structure draft.}
\end{figure}

Whole model structure of NVAE is shown in Fig.(\ref{Structure draft}).

\subsection{Training Strategy}

T2IVAE is trained in a unsupervised way, with a loss function including reconstruction loss and KL regularization term. Reconstruction loss measures the difference between model input and reconstructed output, calculated with the mean square error (MSE) loss as:
\begin{equation}
  L_r(\tilde{X},\hat{X})=\frac{1}{2}\sum_{b,i,j,c=1}^{B,N,N,C}(\tilde{X}_{b,i,j,c}-\hat{X}_{b,i,j,c})^2.
\end{equation}

The purpose of KL regularization term is to make the posterior distribution of latent variables as close to the prior one as possible. Essentially, on the basis of auto-encoder, it introduces randomness and diversity into the encoder output, meanwhile improve the robustness of the decoder. Because of the hierarchical multi-scale encoder-decoder architecture of T2IVAE, KL regularization term is calculated by group and residual Normal distributions are used to improve the calculation efficiency. Specifically, let the prior distribution of the 
\begin{math}
  i^{th}
\end{math} variable of 
\begin{math}
  z_{g}
\end{math} be a Normal distribution:
\begin{displaymath}
  p(z_{g}^{i}|z_{<g}):=\mathcal{N}(\mu_{i}(z_{<g}), \sigma_{i}(z_{<g})).
\end{displaymath}
Then the relative values of prior distribution parameters 
\begin{math}
  \Delta\mu_{i}
\end{math} and 
\begin{math}
  \Delta\sigma_{i}
\end{math} are used to determine the posterior distribution:
\begin{displaymath}
  q(z_{g}^{i}|z_{<g},\tilde{X}):=\mathcal{N}(\mu_{i}(z_{<g})+\Delta\mu_{i}(z_{<g},\tilde{X}), \sigma_{i}(z_{<g})\cdot\Delta\sigma_{i}(z_{<g},\tilde{X})).
\end{displaymath}
For the 
\begin{math}
  g^{th}
\end{math} group, the KL regularization term of each variable is calculated as:
\begin{equation}
  KL(q(z_{g}^{i}|z_{<g},\tilde{X})||p(z_{g}^{i}|z_{<g}))=\frac{1}{2}(\frac{\Delta\mu_{i}^2}{\sigma_{i}^2}+\Delta\sigma_{i}^2-\log\Delta\sigma_{i}^2-1).
\end{equation}
The overall KL regularization term is calculated as:
\begin{equation}
  L_{KL}(\tilde{X},Z)=\sum_{g=1,i=1}^{G,I_{g}}KL(q(z_{g}^{i}|z_{<g},\tilde{X})||p(z_{g}^{i}|z_{<g})),
\end{equation}
where 
\begin{math}
  I_{g}
\end{math} means the scale of latent variables of the 
\begin{math}
  g^{th}
\end{math} group. And the overall loss function is:
\begin{equation}\label{Loss}
  L=L_{r}(\tilde{X},\hat{X})+L_{KL}(\tilde{X},Z).
\end{equation}

The training process based on loss function Eq.(\ref{Loss}) can unsupervisedly fit the time series data, but in the meantime it may fit the anomalous data. Therefore, in order to improve the performance of T2IVAE, we introduce adversarial training to fit the data distribution, making the distribution of reconstructed data as close to that of input data as possible. For the consideration of joint training and model complexity, we make good use of the existing model structure to implement adversarial training inspired by Huang et al\cite{huang2018introvae}.

\begin{algorithm}
  \caption{T2IVAE training}
  \label{T2IVAE training}
  \KwIn{\begin{math}S\end{math}, time series input\;
    \begin{math}N\end{math}, window size\;
    \begin{math}B\end{math}, batch size\;
    \begin{math}epoch\end{math}, number of total training epochs\;
    \begin{math}epoch_{gan}\end{math}, number of adversarial training epochs\;
    \begin{math}\alpha, \beta, m\end{math}, hyper-parameters.
    }
  \KwResult{a trained T2IVAE model with parameters \begin{math}\theta_{Enc}, \theta_{Dec} \end{math}. }
  a set of tensor \begin{math}\check{X} \gets\end{math} preprocess \begin{math}S \end{math}, including standardization, sliding windows and sequence encoding\;
  \begin{math}\theta_{Enc}, \theta_{Dec} \gets\end{math} initialize model parameters\;
  \For{\begin{math}e = 1, \cdots, epoch-epoch_{gan}\end{math}}
  {
    \Repeat{{\rm samples in} \begin{math}\check{X}\end{math} {\rm are enumerated}}{
      \begin{math}\tilde{X}\gets\end{math} random mini-batch from \begin{math}\check{X}\end{math}\;
  \begin{math}Z \gets Enc(\tilde{X})\end{math}\;
  \begin{math}\hat{X} \gets Dec(Z)\end{math}\;
  \begin{math}L_r \gets L_r(\tilde{X},\hat{X})\end{math}\;
  \begin{math}L_{KL} \gets L_{KL}(\tilde{X}, Z)\end{math}\;
  \begin{math}L \gets L_r + L_{KL}\end{math}\;
  perform Adamax updates for \begin{math}\theta_{Enc}, \theta_{Dec} \end{math} based on \begin{math}L\end{math}\;
    }
  }
  \For {\begin{math}e = epoch-epoch_{gan}+1, \cdots, epoch\end{math}}
  {
    \Repeat{{\rm samples in} \begin{math}\check{X}\end{math} {\rm are enumerated}}{
      \begin{math}\tilde{X}\gets\end{math} random mini-batch from \begin{math}\check{X}\end{math}\;
  \begin{math}Z \gets Enc(\tilde{X})\end{math}\;
  \begin{math}Z_p \gets\end{math} samples from prior\;
  \begin{math}\hat{X} \gets Dec(Z), \hat{X}_p \gets Dec(Z_p)\end{math}\;
  \begin{math}L_r \gets L_r(\tilde{X},\hat{X})\end{math}\;
  \begin{math}Z_r \gets Enc(ng(\hat{X})), Z_{pp} \gets Enc(ng(\hat{X}_p))\end{math}\;
  \begin{math}L_d \gets L_r + \beta L_{KL}(\tilde{X}, Z) + \alpha [m-L_{KL}(\hat{X}, Z_r)] + \alpha [m-L_{KL}(\hat{X}_p, Z_{pp})]\end{math}\;
  perform Adamax updates for \begin{math}\theta_{Enc}\end{math} based on \begin{math}L_d\end{math}\;
  
  \begin{math}L_g \gets L_r + \alpha L_{KL}(\hat{X}, Z_r) + \alpha L_{KL}(\hat{X}_p, Z_{pp}) \end{math}\;
  perform Adamax updates for \begin{math}\theta_{Dec}\end{math} based on \begin{math}L_g\end{math}\;
    }
  }
\end{algorithm}

Let the T2IVAE encoder be the discriminator and the T2IVAE decoder be the generator. On one hand, the target of the discriminator is to identify whether the latent variables are encoded from reconstructed data, constructed data from prior or real data, that is, to maximize the 
\begin{math}
  L_{KL}
\end{math} of reconstructed data and constructed data and to minimize the 
\begin{math}
  L_{KL}
\end{math} of real data. On the other hand, the target of the generator is to make the latent variables encoded from constructed data and those from real data are as similar as possible, that is, to minimize the 
\begin{math}
  L_{KL}
\end{math} of constructed data.

We jointly implement VAE training and adversarial training, with loss function of the discriminator and generator:
\begin{equation}\label{LossD}
  L_{d}=L_{r}(\tilde{X}, \hat{X}) + \alpha\sum_{X=\hat{X},\hat{X}_p}[m-L_{KL}(X, Enc(ng(X)))]^+ + \beta L_{KL}(\tilde{X}, Enc(\tilde{X})),
\end{equation}
\begin{equation}\label{LossG}
  L_{g}=L_{r}(\tilde{X}, \hat{X}) + \alpha\sum_{X=\hat{X},\hat{X}_p}L_{KL}(X, Enc(X)),
\end{equation}
where \begin{math}ng(\cdot)\end{math} refers to the saddle point of back propagation of the gradients, \begin{math}\alpha\end{math} and \begin{math}\beta\end{math} are weighting parameters. Because of the instability of adversarial training, we first employ VAE training and then employ adversarial training in the late stage of training process, as stated in Algorithm.(\ref{T2IVAE training}).

\subsection{Detection Strategy}

After model training, we perform anomaly detection through anomaly score estimation, thresholding and pruning. The overall detecting process is stated in Algorithm.(\ref{T2IVAE detecting}).

Firstly, at time step 
\begin{math}
k
\end{math}, a processed tensor
\begin{math}
\check{X}_{k}
\end{math} is input into T2IVAE, and then a reconstructed tensor 
\begin{math}
\hat{X}_{k}
\end{math} is output. The MSE between 
\begin{math}
\check{X}_{k}
\end{math} and 
\begin{math}
\hat{X}_{k}
\end{math} is calculated as the anomaly score at time step 
\begin{math}
k
\end{math}.

Secondly, we use globally adaptive thresholding in order to find anomaly. After estimating anomaly scores at every time step, we calculate the mean and standard deviation of anomaly scores on the whole time, and the threshold is defined as 2 standard deviations above the mean. If the anomaly score at one time step is over the threshold,  an anomaly point is detected, and finally continuous anomaly sequences are obtained after detecting all time steps.

Thirdly, in order to reduce false positives, we introduce a  pruning approach adjusted based on Hundman's method\cite{hundman2018prune}. Given some detected anomaly sequences 
\begin{math}
  \left \{seq_{1},seq_{2},\cdots ,seq_{I} \right \}
\end{math}, we obtain the maximum scores of those sequences 
\begin{math}
  \left \{m_{1},m_{2},\cdots ,m_{I} \right \}
\end{math}, which are then sorted in descending order. After that, we compute the descent rate of the sorted maximums 
\begin{math}
  p_i = (m_{i-1}-m_{i})/m_{i}
\end{math}. When the first sequence matches the condition: 
\begin{displaymath}
  p_i<\theta \ \&\  m_i<4\cdot std  \ \&\    m_i<\lambda\cdot m_1,
\end{displaymath}
the subsequent anomaly sequences are all amended as normal. By default, we set \begin{math}\theta=0.1, \lambda=0.95\end{math}, and \begin{math}std\end{math} refers to the standard deviation of anomaly scores on the whole time series.

Finally, some time sequences are detected as anomalous by T2IVAE.

\begin{algorithm}
  \caption{T2IVAE detecting}
  \label{T2IVAE detecting}
  \KwIn{\begin{math}S\end{math}, time series input with \begin{math}L\end{math} time steps\;
    \begin{math}N\end{math}, window size\;
    \begin{math}B\end{math}, batch size\;
    a trained T2IVAE model with parameters \begin{math}\theta_{Enc}, \theta_{Dec} \end{math}\;
    \begin{math}\theta, \lambda\end{math}, hyper-parameters
    }
  \KwResult{detected anomaly sequences of \begin{math}S\end{math}. }
  a set of tensor \begin{math}\check{X} \gets\end{math} preprocess \begin{math}S \end{math}, including standardization, sliding windows and sequence encoding\;
  \Repeat{{\rm samples in} \begin{math}\check{X}\end{math} {\rm are enumerated}}{
  \begin{math}\tilde{X}\gets\end{math} mini-batch from \begin{math}\check{X}\end{math}\;
  \begin{math}Z \gets Enc(\tilde{X})\end{math}\;
  \begin{math}\hat{X} \gets Dec(Z)\end{math}\;
  \begin{math}\{score_k\} \gets L_r(\check{X}_k, \hat{X}_k)\end{math}, where \begin{math}k = N/2, \cdots, L-N/2\end{math}\;
    }
  \begin{math}mean \gets {\rm mean}(\{score_k\})\end{math}\;
  \begin{math}std \gets {\rm std}(\{score_k\})\end{math}\;
  \For{\begin{math}k = N/2, \cdots, L-N/2\end{math}}{
  \eIf{\begin{math}score_k > mean + 2 \cdot std\end{math}}{
      \begin{math}pred_k = True\end{math}\;
      }{
      \begin{math}pred_k = False\end{math}\;
      }
    }
  \begin{math}\{seq_i\} \gets \{pred_k\}\end{math}, where \begin{math}
  i = 1, \cdots, I\end{math}, and \begin{math}I\end{math} means the number of continuous anomaly sequences\;
  \begin{math}\{m_i\} \gets \{{\rm max\_score}(seq_i)\}\end{math}\;
  \begin{math}\{m_i\} \gets {\rm sort}(\{m_i\}, descending=True)\end{math}\;
  sort \begin{math}\{seq_i\}\end{math} in the same order of \begin{math}\{m_i\}\end{math}\;
  \begin{math}\{p_i\} \gets \{(m_{i-1}-m_{i})/m_{i}\}\end{math}\;
  \For{\begin{math}i = 1, \cdots, I\end{math}}{
  \If{\begin{math}p_i<\theta \ \&\  m_i<4\cdot std  \ \&\    m_i<\lambda\cdot m_1\end{math}}{
      amend \begin{math}\{seq_j\}_{j=i}^I\end{math} as normal\;
      break\;
      }
     }
\end{algorithm}

\section{Experiments}

To evaluate model performance, we have conducted experiments on two public datasets (NASA and Numenta Anomaly Benchmark), and our own dataset collected from network switches. We are going to introduce the experiments specifically in this section.

\subsection{Data}
We use three datasets for model evaluation, including public dataset Numenta Anomaly Benchmark, public dataset NASA, and our own real-world data.

\textbf{Numenta Anomaly Benchmark: }
Numenta Anomaly Benchmark (NAB) provides a public dataset to evaluate anomaly detection algorithms on streaming data. The dataset contains several types of collected time sequences, including artificial data and real-world data\cite{lavin2015evaluating}. We are going to include Artificial (Art), AdExchage (AdEx), AWSCloud(AWS), Traffic(Traf), and Tweets, these five datasets in experiments. Detail of these sub-datasets is available in Table \ref{tablenab}.

\begin{table}
  \caption{Details of public dataset NAB.}
  \label{tablenab}
  \begin{tabular}{ccc}
    \toprule
    Sub-dataset&Sequence num&Sequence length\\
    \midrule
    Art&6&4032\\
    AdEx&5&1624\\
    AWS&17&4032\\
    Traf&7&[1000, 2500]\\
    Tweets&10&15830\\
  \bottomrule
\end{tabular}
\end{table}

\textbf{NASA: }
NASA has provided a public dataset collected from spacecraft telemetry signals, such as radiation intensity, temparature and energy received by sensors. NASA consists of two sub-datasets, Mars Science Laboratory (MSL) and Soil Moisture Active Passive (SMAP). Detail of these sub-datasets is available in Table \ref{tablenasa}.

\begin{table}
  \caption{Details of public dataset NASA.}
  \label{tablenasa}
  \begin{tabular}{ccc}
    \toprule
    Sub-dataset&Sequence num&Sequence length\\
    \midrule
    MSL&27&[1800, 6100]\\
    SMAP&53&[4400, 8700]\\
  \bottomrule
\end{tabular}
\end{table}

\textbf{Our real-world data: }
Our data contain sequences collected from 125 network switches, and with manual label. These sequences indicate volume usage of network equipment, abnormal values may reflect equipment failure, network inaccessibility, network congestion. All sequences are collected from a identical period, with length about 30 days, and the collection interval about 15 min (some series have missing values). Since all data are collected from production environment, it will contain much noise and anomalies not typical, which will test robustness of algorithms.

\subsection{Evaluation Metrics}
In many application scenes in real life, anomalies are usually rare and happen in a form of continuous sequence. As for point anomalies, due to the abruptness and instantaneity, the feasibility and necessity of anomaly detection are questionable. Therefore, we use overlap F1\cite{hundman2018prune} as the evaluation metric. Specifically, if a predicted anomaly sequence overlaps any known anomaly sequence, a true positive is recorded; If a predicted anomaly sequence does not overlap any known anomaly sequence, a false positive is recorded; If a known anomaly sequence does not overlap any predicted anomaly sequence, a false negative is recorded; Lastly, we calculate F1 value with above recorded metrics.

For one dataset, we first obtain the metric of each time series through experiments. Then we calculate the average metric of a sub-dataset. After that, the average metric of a dataset is computed in a weighted way according to the number of time series.

\subsection{Model Setting}
For each time series, we conduct model training and anomaly detection on the whole series. The size of sliding window is 64 and the step size is 1. Since time series data has much less information than image data, a relatively complex model structure is not necessary so we set the number of latent variable groups to 3 and the dimensions of each group to 512, 256 and 128. The number of total training epochs is 50, including 5 adversarial training epochs. The batch size is 128. Adamax optimizer is employed and its learning rate is set to 0.001 for VAE training and 0.0001 for adversarial training. As for adversarial training, we set \begin{math}\alpha\end{math} to 0.005 and \begin{math}\beta\end{math} to 0.1.

\subsection{Results and Analysis}
To conduct our experiments, we have picked VAE-based model DONUT, GAN-based model TADGAN, and statistical model tool Luminol for comparison.

We test \textbf{DONUT} in our experiments for the comparison between DONUT and T2IVAE, since both of them are VAE-based methods. In our paper, results of DONUT are reported from our own experiments using official code, with the best hyper parameters setting after several tests.

\textbf{TADGAN} has multiple variations, using different anomaly score computation methods. In this paper, we report the best method (Critic×DTW) with highest average F1 score on both NAB and NASA taken from original paper, and TADGAN performance on our own real-world data is reported from our own experiments using official code.

\textbf{Luminol}\footnote[4]{Source code is available at: https://github.com/linkedin/luminol} is a open-source tool containing several alternative statistical algorithms, for end-to-end anomaly detection on time series. In our experiments, we uses default Bitmap Algorithm, which breaks time series into chunks and uses the frequency of similar chunks to determine anomaly scores. In this paper, results of Luminol are all from our own experiments with official code.

Firstly, we have made experiments on public datasets NAB and NASA. Since the paper of TADGAN has included a lot of comparison models in experiments on NAB and NASA, we have also listed these results to compare with our model. Specific F1 scores on NAB and NASA for different models are reported in Table \ref{tableresults}.

\begin{table*}
  \caption{Reported F1 scores for model performance comparison on datasets NASA and NAB. Mean score indicates average F1 score of all sequences in the whole dataset, while Total Mean score indicates average score of NASA and NAB. * denotes results taken from the paper of TADGAN. $\rm T2IVAE_{GAN}$ indictes T2IVAE training with GAN strategy.}
  \label{tableresults}
  \begin{tabular}{ccccccccccc}
    \toprule
    &\multicolumn{2}{c}{NASA}&\multicolumn{5}{c}{NAB}&\\
    \cmidrule(lr){2-4}
    \cmidrule(lr){5-10}
    Model&MSL&SMAP&Mean&Art&AdEx&AWS&Traf&Tweets&Mean&Total Mean\\
    \midrule
    MS-$\rm Azure^{*}$&0.218&0.118&0.152&0.125&0.066&0.173&0.166&0.118&0.140&0.146\\
    Luminol&0.278&0.319&0.305&0.121&0.311&0.36&0.225&0.463&0.324&0.315\\
    MAD-$\rm GAN^{*}$&0.111&0.128&0.122&0.324&0.297&0.273&0.412&0.444&0.341&0.232\\
    Dense-$\rm AE^{*}$&0.507&\textbf{0.700}&0.635&0.444&0.267&0.64&0.333&0.057&0.392&0.514\\
    $\rm HTM^{*}$&0.412&0.557&0.508&0.455&0.519&0.571&0.474&0.526&0.525&0.516\\
    LSTM-$\rm AE^{*}$&0.507&0.672&0.616&0.545&0.571&\textbf{0.764}&0.552&0.542&0.630&0.623\\
    $\rm DeepAR^{*}$&0.583&0.453&0.497&0.545&0.615&0.390&0.600&0.542&0.505&0.501\\
    $\rm Arima^{*}$&0.492&0.420&0.444&0.353&0.583&0.518&0.571&0.567&0.524&0.484\\
    $\rm LSTM^{*}$&0.460&0.690&0.612&0.375&0.538&0.474&\textbf{0.634}&0.543&0.509&0.561\\
    DONUT&0.511&0.533&0.526&0.354&0.474&0.491&0.059&0.591&0.427&0.476\\
    $\rm TADGAN^{*}$&\textbf{0.623}&0.680&\textbf{0.661}&\textbf{0.667}&\textbf{0.667}&0.610&0.455&0.605&0.600&0.630\\
    \midrule
    T2IVAE&0.568&0.672&0.637&0.613&0.556&0.673&0.476&\textbf{0.639}&0.613&0.625\\
    $\rm T2IVAE_{GAN}$&0.595&0.679&0.651&0.626&0.572&0.692&0.595&0.628&\textbf{0.639}&\textbf{0.645}\\
    \bottomrule
  \end{tabular}
\end{table*}

\textbf{ARIMA}, \textbf{LSTM}, \textbf{HTM} are three prediction-based models tested and reported in the paper of TADGAN\cite{geiger2020tadgan}. Prediction-based methods learn history of time series and make prediction for next several time steps, and real observations with prediction error above threshold will be detected as anomalies.

\textbf{LSTM AE}, \textbf{Dense AE}, and \textbf{MAD-GAN} are reconstruction-based models tested and reported in the paper of TADGAN\cite{geiger2020tadgan}. LSTM AE and Dense AE are both based on standard Auto-Encoders but with different encoding layers (LSTM layers or Dense layers), while MAD-GAN\cite{li2019mad} is based on GAN to train the model for reconstruction. For anomaly detection, reconstruction error is used to determine the observation is abnormal or not.

\textbf{MS Azure} and \textbf{DeepAR} are two commercial tools reported in the paper of TADGAN\cite{geiger2020tadgan}. MS Azure from Microsoft uses Spectral Residual Convolutional Neural Networks (SR-CNN) for anomaly detection, which applies Spectral Residual for sequence decomposition and applies Convolutional Neural Networks as supervised learning model for detecting anomalies. DeepAR is a probabilistic forecasting model with autoregressive recurrent networks from Amazon, which can be used as a prediction-based model, and let prediction error to detect anomalies.

According to Table 3, T2IVAE (with GAN training strategy) performs best (F1 score is 0.639) on NAB, and performs second best (F1 score is 0.651) on NASA with small disadvantage compared to TADGAN, which makes it the best model considering average F1 score (0.645) of NASA and NAB. Furthermore, we can have three conclusions about the performance of T2IVAE:

First, T2IVAE and DONUT are both VAE-based algorithms, but T2IVAE performs better on both NASA and NAB. Compared with DONUT, T2IVAE replaces classical VAE structure with more complicated structure as NVAE, and transform time series into 2D image to present temporal correlation before input. The improved performance proves, better-designed VAE structure, and more information on temporal correlation, are effective for time series anomaly detection.

Second, compared to other reconstruction-based models, our model with GAN training strategy still performs better (except disadvantage compared to TADGAN on NASA). This shows the effectiveness of VAE as the backbone of reconstruction-based model structure.

Third, we have tested both T2IVAE and $\rm T2IVAE_{GAN}$ on public datasets. The only difference between these two models, is that $\rm T2IVAE_{GAN}$ applies GAN training strategy in last several training epochs. In our experiments, we replace last 5 training epochs with GAN strategy. Compare the performance of these two models, we can find such improvement can help improve performance on anomaly detection.

On our real-world data, we pick Luminol, DONUT, TADGAN (performs best on NAB and NASA except our model) for comparison. Results are shown in Table \ref{tabproj}.

\begin{table}
  \caption{Reported F1 scores for model performance comparison on our real-world dataset. $\rm T2IVAE_{GAN}$ indicates T2IVAE training with GAN strategy.}
  \label{tabproj}
  \begin{tabular}{cc}
    \toprule
    Model&F1 Score\\
    \midrule
    Luminol&0.315\\
    DONUT&0.481\\
    TADGAN&0.424\\
    \midrule
    T2IVAE&0.494\\
    $\rm T2IVAE_{GAN}$&\textbf{0.504}\\
    \bottomrule
  \end{tabular}
\end{table}

On our real-world data, except our models, DONUT performs best this time, while TADGAN performs best on NASA and NAB, which means models may be suitable for different datasets. Both T2IVAE and $\rm T2IVAE_{GAN}$ perform better than any other comparison models, and $\rm T2IVAE_{GAN}$ performs better than T2IVAE again, which indicates the effectiveness of improvements made for T2IVAE (including sequence encoding, NVAE structure, and GAN training strategy).

\section{Conclusion}
In this paper, we focus on univariate time series anomaly detection and propose T2IVAE, an unsupervised learning model based on NVAE. T2IVAE encode original time series into 2D image to extract correlation feature, apply multiple VAE sampling procedures, and use reconstruction loss to detect time steps with abnormal values in time series. Furthermore, we’ve also explored to optimize model performance with GAN training strategy and adaptive anomaly threshold. To evaluate model and compare it with other algorithms, we conduct on public datasets and our own collected real-world dataset. T2IVAE has achieved 0.639 F1 score on NAB dataset, 0.651 F1 score on NASA dataset, and 0.504 F1 score on our real-world dataset, making it outperform other comparison models. This paper mainly shows that, better performing VAE architecture and better designed feature engineering, can help improve performance on time series anomaly detection.


\bibliographystyle{ACM-Reference-Format}
\bibliography{paperarxiv}


\end{document}